\documentclass[11pt,a4paper]{article}
\usepackage[hyperref]{acl2020}
\usepackage{times}
\usepackage{latexsym}

\usepackage{microtype}

\usepackage{amsmath}
\usepackage{graphicx}
\usepackage{arydshln}
\usepackage{multirow}
\usepackage{subcaption}

\aclfinalcopy 
\def\aclpaperid{83} 

\DeclareMathOperator*{\argmax}{argmax}

\title{Beyond Domain APIs: Task-oriented Conversational Modeling with Unstructured Knowledge Access}

\author{
  Seokhwan Kim, Mihail Eric, Karthik Gopalakrishnan,\\
  \bf{Behnam Hedayatnia, Yang Liu, Dilek Hakkani-Tur}\\
  Amazon Alexa AI, Sunnyvale, CA, USA\\
  \{seokhwk,mihaeric,karthgop,behnam,yangliud,hakkanit\}@amazon.com\\
  }

\date{}

\begin{document}
\maketitle
\begin{abstract}
  Most prior work on task-oriented dialogue systems are restricted to a limited coverage of domain APIs, while users oftentimes have domain related requests that are not covered by the APIs.
  In this paper, we propose to expand coverage of task-oriented dialogue systems by incorporating external unstructured knowledge sources.
  We define three sub-tasks: knowledge-seeking turn detection, knowledge selection, and knowledge-grounded response generation,
  which can be modeled individually or jointly.
  We introduce an augmented version of MultiWOZ 2.1, which includes new out-of-API-coverage turns and responses grounded on external knowledge sources.
  We present baselines for each sub-task using both conventional and neural approaches.
  Our experimental results demonstrate the need for further research in this direction to enable more informative conversational systems.
\end{abstract}

\section{Introduction}
\label{sec:intro}
Traditionally, task-oriented dialogue systems have focused on providing information and performing actions that can be handled only by given databases or APIs.
However, in addition to task-focused requests, users also have needs that go beyond what is provided by the backend resources.
For example, while most virtual assistants can help users book a hotel, a restaurant or movie tickets, they fall short of answering potential follow-up questions users may have,
such as: where to park vehicles; whether they are allowed to bring pets or children to the reserved place; or what the cancellation policy is.
No API/DB entry is usually available to handle such requests.
On the other hand, relevant domain knowledge is already available on web pages in the form of descriptions, FAQs and customer reviews for many of these out-of-coverage scenarios.
Since current dialogue systems don't incorporate these external knowledge sources into task-oriented conversational modeling,
users need to visit the websites by themselves to find out any additional information beyond API/DB coverage, making conversational interactions inefficient.

In this work, we propose a new conversational modeling task towards frictionless task-oriented scenarios,
where the flow of the conversation does not break when users have requests that are out of the coverage of APIs/DB but potentially are already available in external knowledge sources.
Inspired by recent studies on knowledge-grounded conversational modeling~\cite{zhou2018commonsense,dinan2018wizard,galley2019grounded,gopalakrishnan2019topical},
our proposed task aims to develop end-to-end dialogue systems to understand relevant domain knowledge, and generate system responses with the selected knowledge.
Different from previous work on social conversations~\cite{ritter2011data,vinyals2015neural,serban2017hierarchical}, this task addresses task-oriented conversations grounded on fine-grained domain-level or entity-level knowledge sources related to given dialogue contexts.

Figure~\ref{fig:example} shows an example conversation with unstructured knowledge access.
The user utterances at turns $t=\{3, 7\}$ and $t=\{11, 15\}$ request the policy details about bringing pets and making payments, respectively, which are out of the coverage of the structured domain APIs.
On the other hand, the relevant knowledge contents can be found from the external sources as in the rightmost column which includes the QA snippets from the FAQ lists for each corresponding entity within domains such as train, hotel, or restaurant.
With access to these unstructured external knowledge sources, the agent managed to continue the conversation with no friction by responding adequately at the turns $t=\{4, 8, 12, 16\}$.

\begin{figure*}[t]
\scriptsize
\begin{tabular}{r l p{7.5cm} p{5.5cm}}
  $t$ & Speaker & Utterance ($u_t$) & Relevant Knowledge Snippets from FAQs\\ \hline
1 & User & I need a train from King's Lynn to Cambridge arriving by 17:45 on Sunday. \\
2 & Agent & I have train TR6003 leaving cambridge at 5:11 and arriving at 5:58 on sunday \\ \hdashline[.4pt/1pt]
3 & User & I also need to bring my dog. Do they allow pets? & \multirow{2}{5.5cm}{Q: Can I bring my pets to trains? \newline A: We happily welcome dogs and cats up to 20 pounds.}\\
4 & Agent & \textbf{Yes, you can travel with your dog up to 20 pounds.} \\ \hdashline[.4pt/1pt]
5 & User & That sounds great. I also need help finding a place to stay in the South.\\
6 & Agent & I have 2 options available. The Lensfield Hotel and Gonville Hotel. \\ \hdashline[.4pt/1pt]
7 & User & Do either of them allow to stay with my dog? & \multirow{2}{5.5cm}{Q1: Can I bring my dog to Gonville Hotel? \newline A1: Pets are permitted. But charges may be applicable.}\\
8 & Agent & \textbf{You could stay with your dog at Gonville with a fee.} \\ \cdashline{4-4}[.4pt/1pt]
    & &  \textbf{Would you like me to book it?} & \multirow{2}{5.5cm}{Q2: Can I bring my dog? \newline A2: Pets are not allowed at the Lensfield Hotel.}\\
  \\
\hdashline[.4pt/1pt]
9 & User & Maybe later. How about a place to eat nearby? Chinese food would be great. \\
10 & Agent & The peking restaurant is a nice place. Do you need reservations? \\ \hdashline[.4pt/1pt]
11 & User & Before that, could you confirm that this restaurant accepts AMEX? & \multirow{2}{5.5cm}{Q: What type of payments are accepted? \newline A: Peking Restaurant accepts cash only.}\\
12 & Agent & \textbf{Unfortunately, the Peking restaurant accepts cash only. Would it work?} \\ \hdashline[.4pt/1pt]
13 & User & Okay. Can you book a table for 4 at 18:30 on Monday, please? \\
14 & Agent & Booking was successful. Do you have any other question?\\ \hdashline[.4pt/1pt]
15 & User & What about the hotel? Can I use my credit card there? & \multirow{2}{5.5cm}{Q: What credit cards are accepted by Gonville Hotel? \newline A: You can use AMEX, Visa and Mastercard.}\\
16 & Agent & \textbf{Yes, Gonville Hotel accepts all major credit cards including AMEX.} \\
\end{tabular}
\caption{Examples of task-oriented conversations with unstructured knowledge access}
\label{fig:example}
\end{figure*}

\begin{figure*}[t]
\centering
\includegraphics[width=\textwidth]{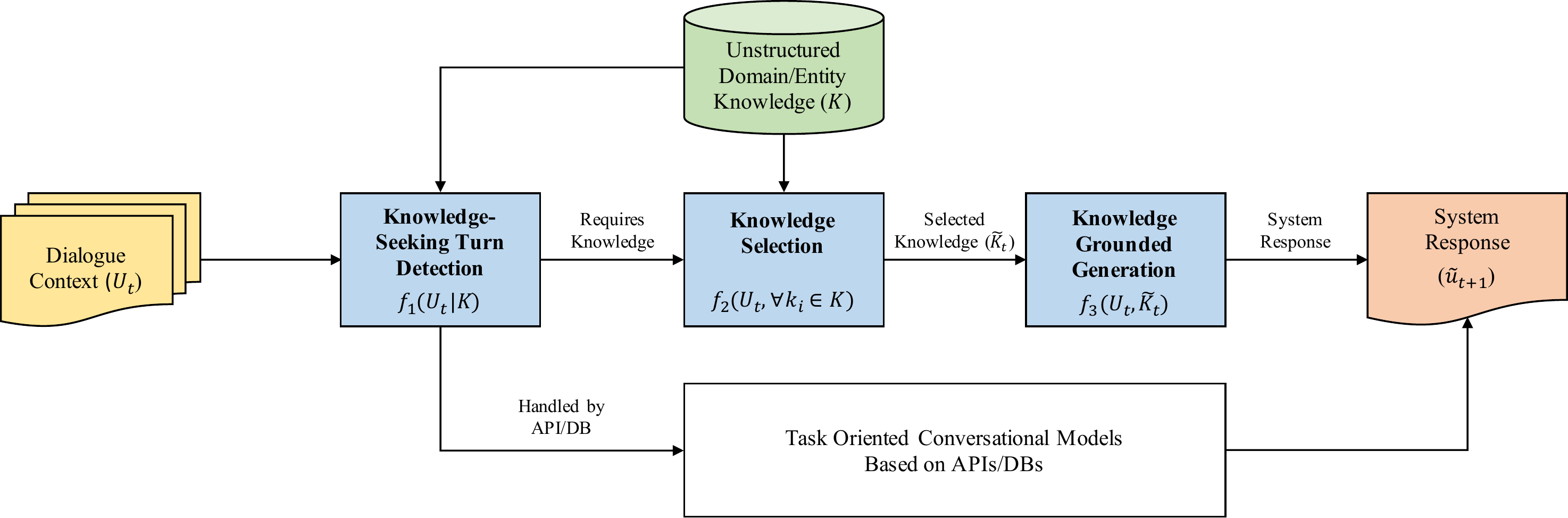}
\caption{A baseline architecture for task-oriented conversational modeling grounded on unstructured knowledge}
\label{fig:overview}
\end{figure*}

\section{Related Work}
\label{sec:related}

Task-oriented dialogue systems aim to enable users to complete tasks by interacting with an automated agent in natural language~\cite{young2013pomdp}.
These systems typically convert user utterances to a semantic representation (such as domain, intent, and slots~\cite{tur2011spoken}) based on what is used by the backend resources (such as APIs) that accomplish the tasks.
At each turn, the dialogue system decides the next action to take based on the estimated dialogue state as well as any results or responses from the backend resources~\cite{levin2000stochastic,singh2002optimizing,williams2007partially}.
The next action, which is typically in the form of a semantic frame formed of dialogue acts, arguments and values, is converted to a natural language response to the user by natural language generation~\cite{perera2017recent}.

On the other hand, social conversational systems typically follow an end-to-end approach, and aim to generate target responses based on the previous conversation context~\cite{ritter2011data,vinyals2015neural,serban2017hierarchical}.
Ghazvininejad et al.~\shortcite{ghazvininejad2018knowledge} proposed an extension to these models that grounds the responses on unstructured, textual knowledge, by using end-to-end memory networks where an attention over the knowledge relevant to the conversation context is estimated.
Along similar lines, Liu et al.~\shortcite{liu2018knowledge} used pattern matching, named entity recognition and linking to find facts relevant to the current dialogue and other related entities from a knowledge base.
Zhou et al.~\shortcite{zhou2018commonsense} proposed both static and dynamic graph attention mechanisms for knowledge selection and response generation, respectively, using knowledge graphs.
More recently, Dinan et al.~\shortcite{dinan2018wizard} and Gopalakrishnan et al.~\shortcite{gopalakrishnan2019topical} both have publicly released large conversational data sets, where knowledge sentences related to each conversation turn are annotated.
Our proposed task, data, and baseline models in this work differ from these studies in the following aspects: we target task-oriented conversations with more clear goals and explicit dialogue states than social conversations;
and we aim to incorporate task-specific domain knowledge instead of commonsense knowledge.

The other line of related work is machine reading comprehension which aims to answer questions given unstructured text~\cite{richardson2013mctest,hermann2015teaching,rajpurkar2016squad} and has later been extended to conversational question answering~\cite{choi2018quac,reddy2019coqa}.
In our work, the document required to generate a response needs to be identified according to the conversation context.
The responses are also different in that, rather than plain answers to factual questions, we aim to form factually accurate responses that seamlessly blend into the conversation.

\section{Problem Definition}
\label{sec:problem}

We define an unstructured knowledge-grounded task-oriented conversational modeling task based on a simple baseline architecture (Figure~\ref{fig:overview})
which decouples turns that could be handled by existing task-oriented conversational models with no extra knowledge
and turns that require external knowledge resources. 
In this work, we assume that a conventional API-based system already exists and focus on the new knowledge access branch which takes a dialogue context $U_t = \{u_{t-w+1}, \cdots, u_{t-1}, u_t\}$ and knowledge snippets $K = \{k_1, \cdots, k_n\}$,
where $u_{i}$ is the $i$-th utterance in a given dialogue, $t$ is the time-step of the current user utterance to be processed, $w$ is the dialogue context window size. 

Our proposed task aims to generate a context-appropriate system response $\tilde{u}_{t+1}$ grounded on a set of relevant knowledge snippets $\tilde{K} \subset K$.
The remainder of this section presents the detailed formulations of the following three sub-tasks: \textit{`Knowledge-seeking Turn Detection'}, \textit{`Knowledge Selection'}, and \textit{`Knowledge-grounded Response Generation'}.

\subsection{Knowledge-seeking Turn Detection}
\label{sec:problem:task1}

For each given turn at $t$, a system first needs to decide whether to continue an existing API-based scenario or trigger the knowledge access branch.
We call this task \textit{Knowledge-seeking Turn Detection}.
This problem is defined as a binary classification task formulated as follows:
\begin{equation*}
  f_1(U_t|K) = \left\{ \begin{array}{ll}
                       1 & \mbox{if $\exists k \in K$ satisfies $u_t$,}\\
                       0 & \mbox{otherwise,}
                     \end{array} \right.
\end{equation*}
which we assume that every turn can be handled by either branch in this work.
For the examples in Figure~\ref{fig:example}, $f_1(U_t|K)=1$ for the knowledge-seeking turns at $t=\{3, 7, 11, 15\}$, while $f_1(U_t|K)=0$ for the other user turns at $t=\{1, 5, 9, 13\}$.

\subsection{Knowledge Selection}
\label{sec:problem:task2}

Once a given user turn at $t$ is determined as a knowledge-seeking turn by $f_1(U_t|K)$,
it moves forward with \textit{Knowledge Selection} to sort out the relevant knowledge snippets.
This task takes each pair of $U_t$ and $k_i \in K$ and predicts whether they are relevant or not as follows:
\begin{equation*}
  f_2(U_t, k_i) = \left\{ \begin{array}{ll}
                         1 & \mbox{if $k_i \in K$ is relevant to $U_t$,}\\
                         0 & \mbox{otherwise.}
                       \end{array} \right.
\end{equation*}

Different from other information retrieval problems taking only a short single query, this knowledge selection task must be highly aware of the dialogue context.
For example, $u_3$ and $u_7$ themselves in Figure~\ref{fig:example} share the same question type with similar surface form,
but the relevant knowledge snippets would vary depending on their dialogue states across different domains.
Even within a single domain, fine-grained dialogue context needs to be taken into account to select proper knowledge snippets corresponding to a specific entity,
for example, `Peking Restaurant' and `Gonville Hotel' for $u_{11}$ and $u_{15}$ against any other restaurants and hotels, respectively.

Since more than one knowledge snippet can be relevant to a single turn, as for $u_7$ in Figure~\ref{fig:example},
we form a task output $\tilde{K}$ including all the positive knowledge snippets from $f_2(U_t, k)$, as follows:
\begin{equation*}
  \tilde{K_t} = \left\{k_i | k_i \in K \wedge f_2(U_t, k_i) = 1 \right\} \subset K.
\end{equation*}

\subsection{Knowledge-grounded Generation}
\label{sec:problem:task3}

Finally, a system response $\tilde{u}_{t+1}$ is generated based on both dialogue context $U_t$ and the selected knowledge snippets $\tilde{K_t}$, as follows:
\begin{equation*}
  f_3(U_t, \tilde{K_t}) = \tilde{u}_{t+1}.
\end{equation*}
Each generated response is supposed to provide the user with the requested information grounded on the properly selected knowledge sources.
In addition, the response should be naturally connected to the previous turns.
The knowledge-grounded responses in Figure~\ref{fig:example} focus not only on delivery of the information by knowledge access, but also maintain natural conversation.
For example, the responses at $t=\{4,8\}$ paraphrase written sentences into a colloquial style, the responses at $t=\{4,16\}$ acknowledge before giving a statements, the responses at $t=\{8,12\}$ ask a follow-up question to the user.

\begin{figure*}
  \centering
  \begin{subfigure}[b]{0.32\textwidth}
    \includegraphics[width=\textwidth]{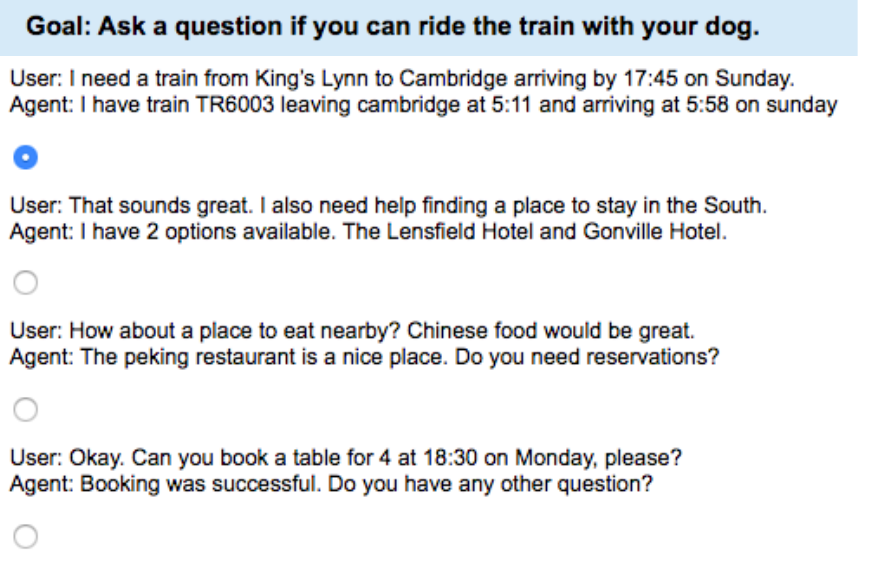}
    \caption{Positions for augmentation}
    \label{fig:mturk_task1}
  \end{subfigure}
  ~
  \begin{subfigure}[b]{0.32\textwidth}
    \includegraphics[width=\textwidth]{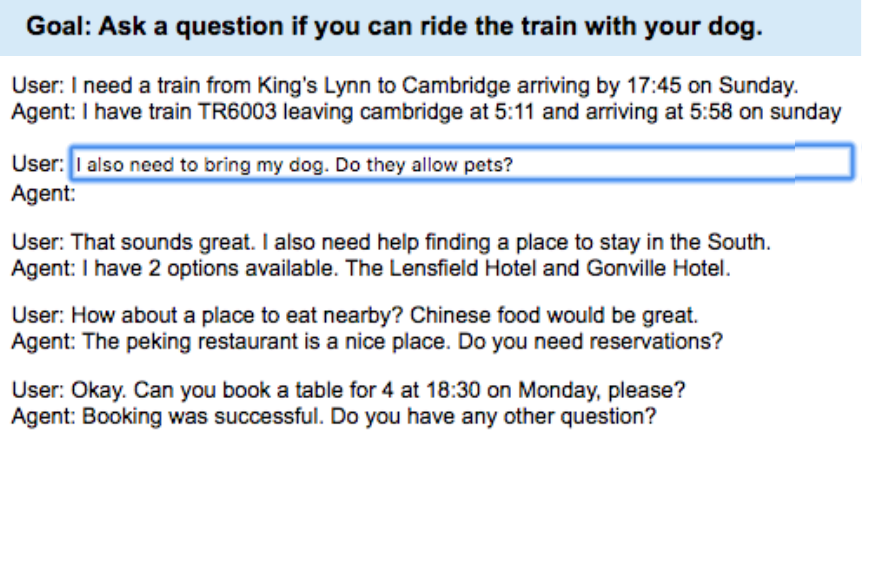}
    \caption{User utterances}
    \label{fig:mturk_task2}
  \end{subfigure}
  ~
  \begin{subfigure}[b]{0.32\textwidth}
    \includegraphics[width=\textwidth]{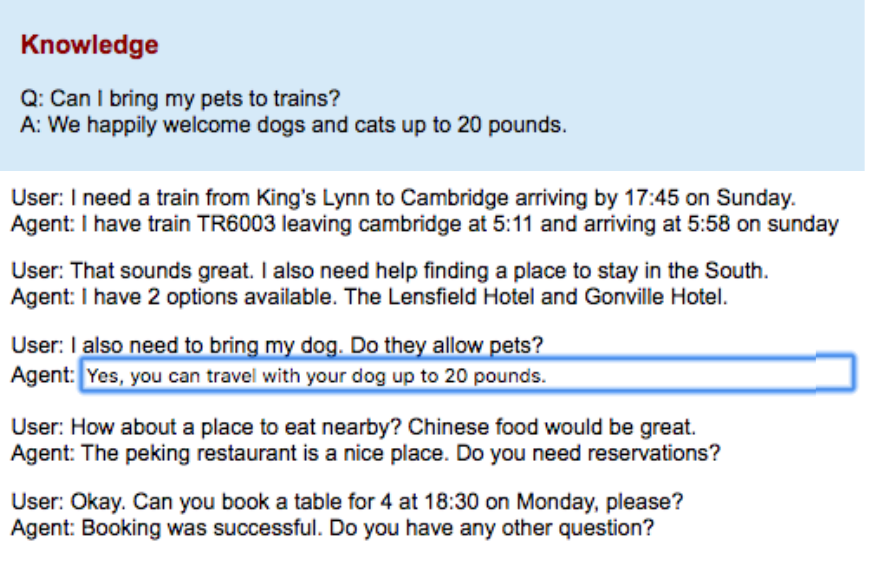}
    \caption{System responses}
    \label{fig:mturk_task3}
  \end{subfigure}
  \caption{Crowdsourcing user interfaces for MultiWOZ data augmentation with knowledge access turns}\label{fig:mturk_tasks}
\end{figure*}

\section{Data}
\label{sec:data}


To address the proposed research problems, we collected an augmented version of MultiWOZ 2.1~\cite{budzianowski2018multiwoz,eric2019multiwoz} with out-of-API-coverage turns grounded on external knowledge sources beyond the original database entries.
This was incrementally done by the following three crowdsourcing tasks.

First, crowd workers were given a dialogue sampled from the original MultiWOZ 2.1 conversations
and asked to indicate an appropriate position to insert a new turn about a selected subject from external knowledge categories (Figure~\ref{fig:mturk_task1}).
This task aims to collect user behaviors about when to ask a knowledge-seeking question for a given subject.
It corresponds to the knowledge-seeking turn detection sub-task in Section~\ref{sec:problem:task1}.

Then, they were asked to write down a new user utterance at each selected position in the first task to discuss about a given corresponding subject (Figure~\ref{fig:mturk_task2}),
which is for both knowledge-seeking turn detection (Section~\ref{sec:problem:task1}) and knowledge selection (Section~\ref{sec:problem:task2}) sub-tasks.
In order to collect various expressions, a single task with the same dialogue context and knowledge category was assigned to multiple crowd workers in parallel.

Finally, we collected the agent's response to each question collected in the previous step.
In this task (Figure~\ref{fig:mturk_task3}), crowd workers were given external knowledge sources for each category and asked to convert them into a system response which is more colloquial and coherent to both the question and dialogue context.
This task aims at knowledge-grounded response generation (Section~\ref{sec:problem:task3}).

Our proposed pipeline for data collection has the following advantages over Wizard-of-Oz (WoZ) approaches.
First, it is more efficient and scalable, since every task can be done by a single crowd worker independently from others, while WoZ requires to pair up two crowd workers in real time.
This aspect enables us to have more control in the whole process compared to the end-to-end data collection entirely by crowd workers from scratch.
Furthermore, the intermediate outcomes from each phase can be utilized to build conversational models with no additional annotation.

\begin{table}
  \centering
  \small
  \begin{tabular}{l r r r r}
    Split & \# dialogues & \# augmented turns & \# utterances\\ \hline
    Train & 8,438 & 7,169 & 127,894 \\
    Valid & 1,000 & 923 & 16,594 \\
    Test  & 1,000 & 980 & 16,704 \\ \hdashline[.4pt/1pt]
    Total & 10,438 & 9,072 & 161,192
  \end{tabular}
  \caption{Statistics of the data divided into training, validation, and test purposes. The total number of utterances includes both the original and augmented turns.}
  \label{tbl:data_stats}
\end{table}

\begin{table}
  \centering
  \small
  \begin{tabular}{l r r r r}
    & Domain-level & & \multicolumn{2}{c}{Entity-level} \\
    Domain & \# snippets & & \# entities & \# snippets \\ \hline
    Hotel & 24 & & 27 & 477 \\
    Restaurant & 8 & & 81 & 401 \\
    Train & 20 & & - & - \\
    Taxi & 8 & & - & - \\ \hdashline[.4pt/1pt]
    Total & 60 & & 108 & 878
  \end{tabular}
  \caption{Statistics of domain-/entity-level knowledge snippets collected from FAQ webpages}
  \label{tbl:knowledge_stats}
\end{table}

\begin{figure}[t]
  \includegraphics[width=\linewidth]{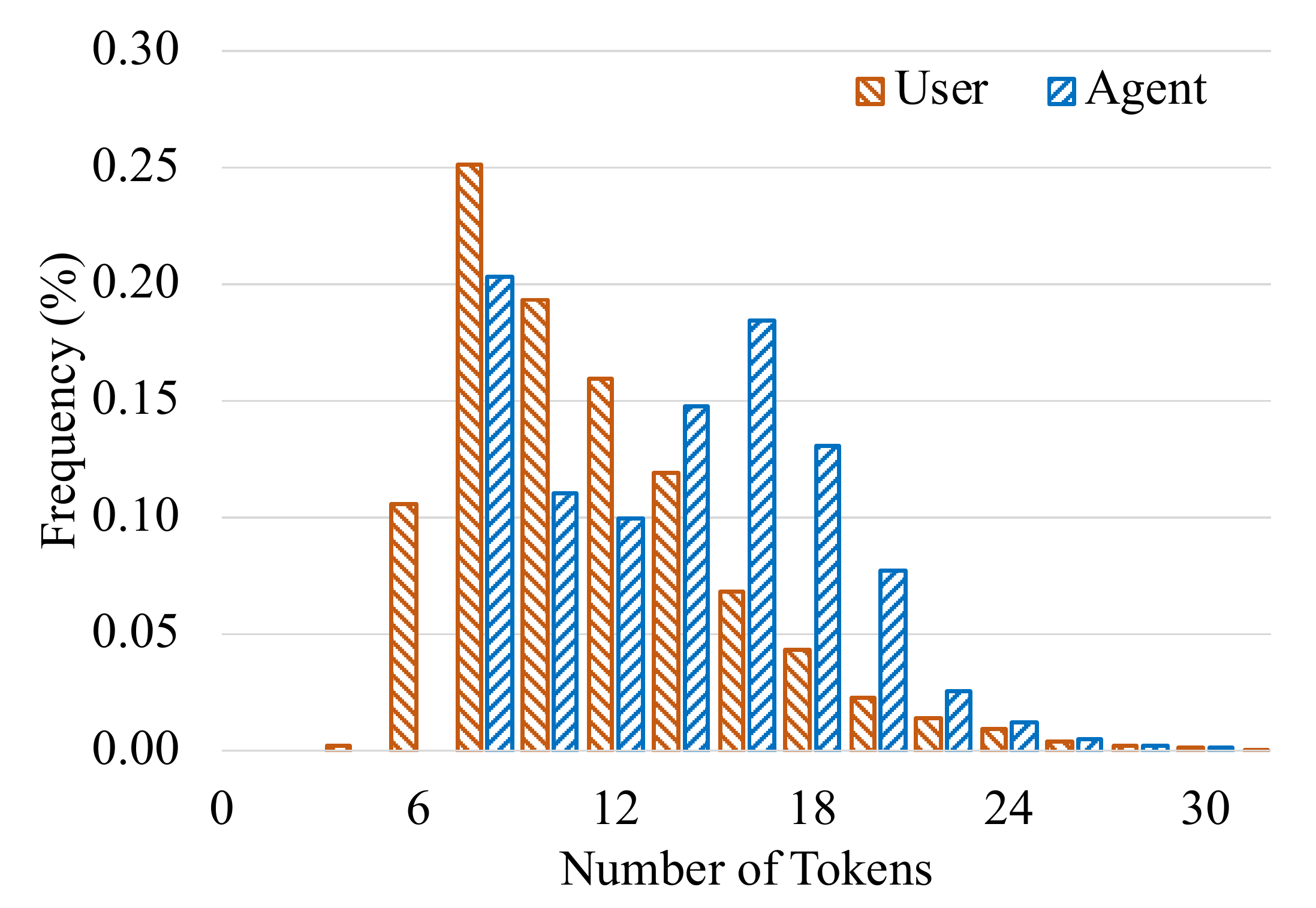}
  \caption{Distribution of number of tokens of the augmented utterances}
  \label{fig:length_dist}
\end{figure}

\begin{figure}[t]
  \includegraphics[width=\linewidth]{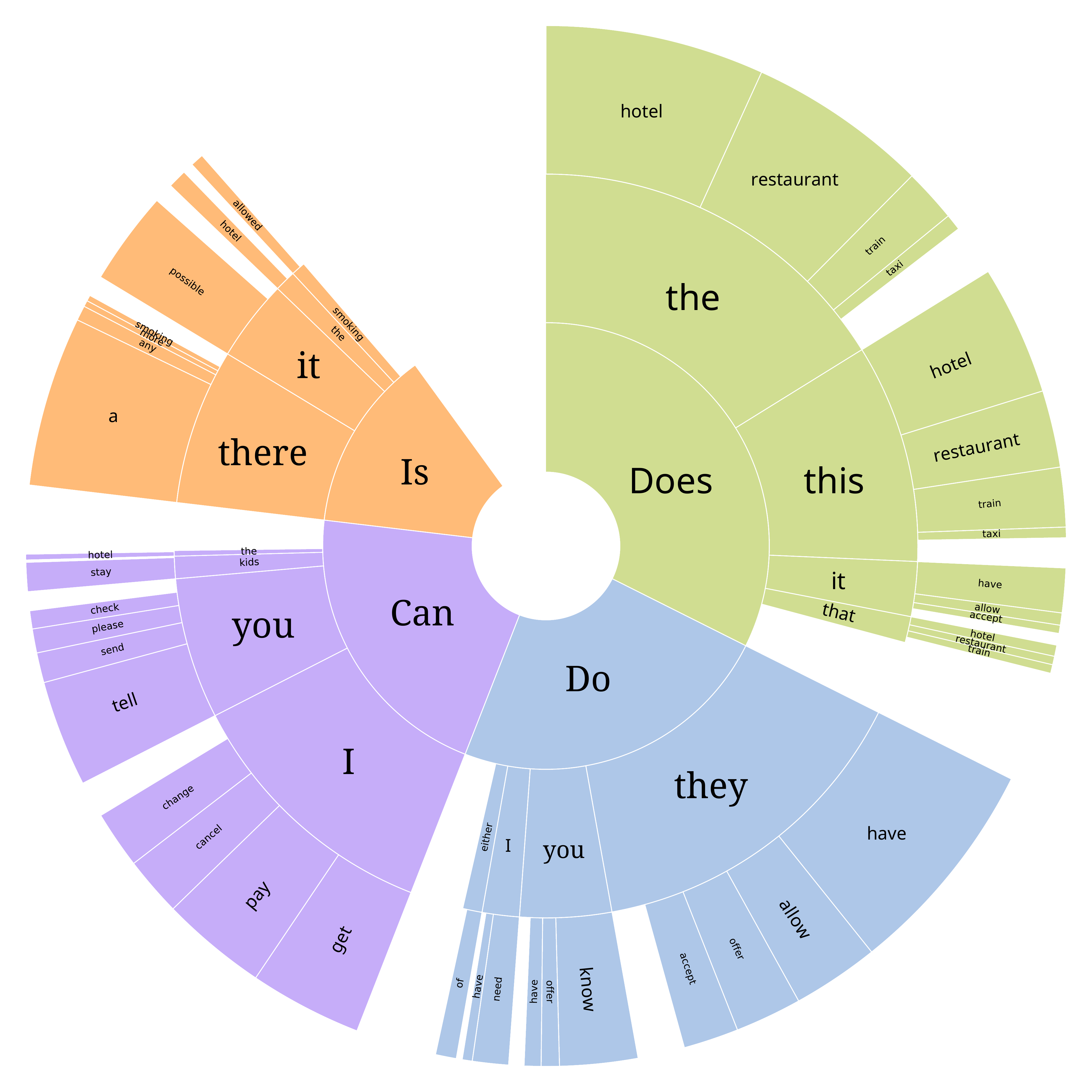}
  \caption{Distribution of trigram prefixes of the augmented user utterances}
  \label{fig:prefix}
\end{figure}

Table~\ref{tbl:data_stats} shows the statistics of the collected data sets.
A total of 9,072 utterance pairs are newly collected in addition to the original MultiWOZ dialogues,
each of which is linked to corresponding knowledge snippets among 938 question-answer pairs (Table~\ref{tbl:knowledge_stats})
collected from the FAQ webpages about the domains and the entities in MultiWOZ databases.
Figure~\ref{fig:length_dist} shows the length distribution of the augmented utterances.
Similar to the original MultiWOZ~\cite{budzianowski2018multiwoz}, the agent responses are longer than the user utterances, which have 12.45 and 9.85 tokens on average spoken by agents and users, respectively.
Figure~\ref{fig:prefix} presents the distribution of trigram prefixes of the augmented user utterances with various types of follow-up questions that go beyond the coverage of domain APIs.

\section{Methods}
\label{sec:methods}

In this section, we present baseline methods for the problems defined in Section~\ref{sec:problem}.
Specifically, we introduce both a non-machine learning approach and a neural baseline model for each sub-task.


\subsection{Knowledge-seeking Turn Detection}
\label{sec:methods:task1}
For the knowledge-seeking turn detection, we compare two baselines with unsupervised anomaly detection and supervised classification methods.

\subsubsection{Unsupervised Anomaly Detection}
\label{sec:methods:task1:anomaly}

In the first baseline, we consider the task as an anomaly detection problem that aims to identify the turns that are out of the coverage of conventional API-based requests.
Given the assumption that there is no knowledge-seeking turn available in most task-oriented dialogue data,
we applied an unsupervised anomaly detection algorithm, Local Outlier Factor (LOF)~\cite{breunig2000lof}.
The algorithm compares the local densities between a given input instance and its nearest neighbors.
If the input has a significantly lower density than the neighbors, it is considered an anomaly.

We built a knowledge-seeking turn detector with the LOF implementation in PyOD~\cite{zhao2019pyod} with its default configurations.
The system includes all the user utterances in the original MultiWOZ 2.1 training set.
Every utterance in both training and test sets was encoded by the uncased pre-trained BERT~\cite{Devlin2019BERTPO} model.

\subsubsection{Neural Utterance Classification}
\label{sec:methods:task1:classification}
If training data is available for the knowledge-seeking turn detection,
the most straightforward solution will be training a binary classifier in a supervised manner.
In this experiment, we fine-tuned the uncased pre-trained BERT~\cite{Devlin2019BERTPO} model on the training data in Section~\ref{sec:data}.
The model takes each single user utterance $u_t$ as an input and generates the utterance representation as the final layer output for $[CLS]$ which is a special token in the beginning of the input sequence.
We added a single layer feedforward network on top of the utterance embeddings, which was trained with binary cross-entropy loss for three epochs.
We used a mini-batch size of 128 with truncated utterances up to 256 tokens.

\subsection{Knowledge Selection}
\label{sec:methods:task2}
In our experiments, we consider two variants of the knowledge selector: unsupervised knowledge-retrieval baselines and supervised neural Transformer architectures.

\begin{figure}[t]
  \includegraphics[width=\linewidth]{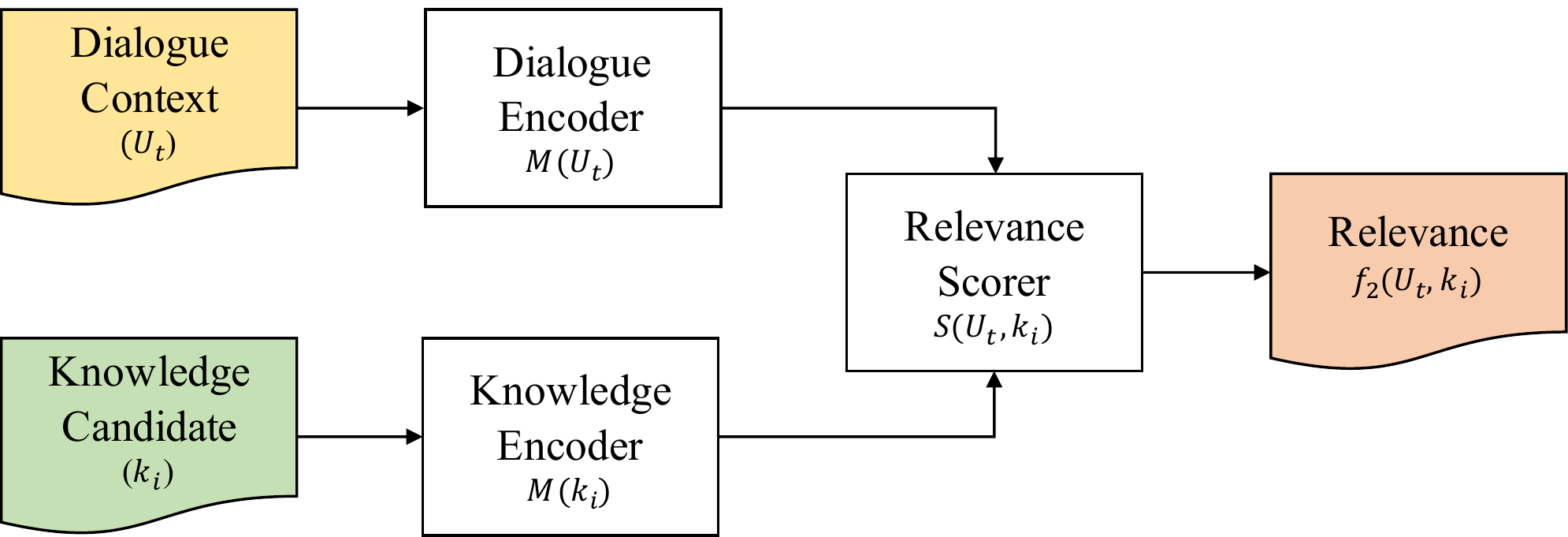}
  \caption{Retrieval baseline for knowledge selection}
  \label{fig:baseline_task2_ir}
\end{figure}

\begin{figure}[t]
  \includegraphics[width=\linewidth]{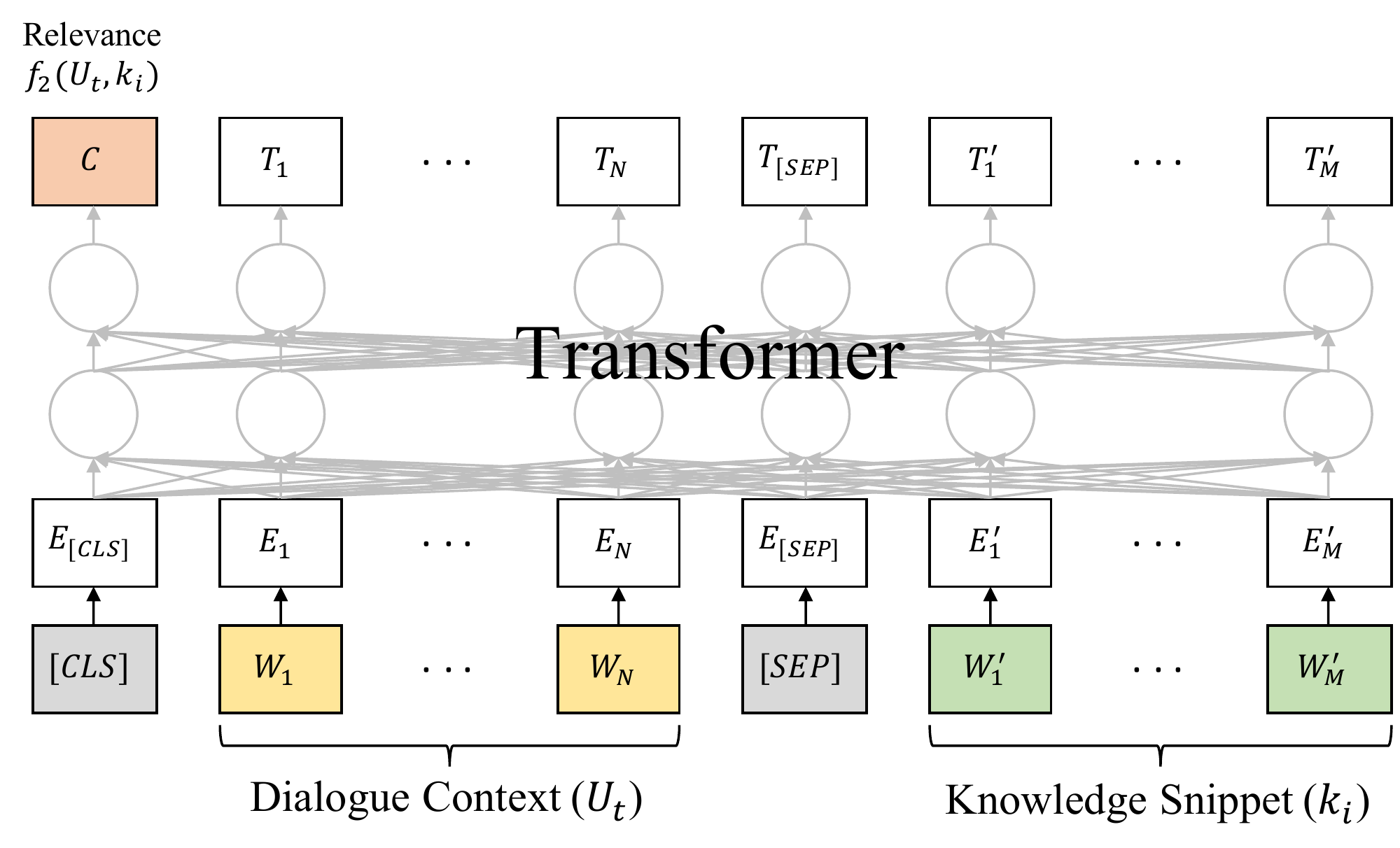}
  \caption{BERT-based knowledge selection baseline}
  \label{fig:baseline_task2_bert}
\end{figure}

\subsubsection{Unsupervised Knowledge Retrieval}
\label{sec:methods:task2:retrieval}

First, we propose the unsupervised knowledge selection baselines using information retrieval (IR) algorithms (Figure~\ref{fig:baseline_task2_ir}).
Let us denote an encoder function $M$ mapping the concatenation of all the sentences in a query or a document to a fixed-dimensional weight vector.
In this work, we take the dialogue context $U_t$ as a query and each knowledge snippet $k_i$ as a candidate document.
When scoring entity-level knowledge, we also add the name of the entity to each document $k_i$ being scored as this helps differentiate among potentially ambiguous knowledge contents that may be applicable to multiple entities.

Our IR model then computes the following cosine similarity score per knowledge snippet:
\begin{equation*}
  S\left(U_t, k_i\right) = \cos\left(M(U_t), M(k_i)\right),
\end{equation*}
where we finally take the most relevant document as a selected knowledge in the following fashion:
\begin{equation*}
  f_2(U_t, k_i) = \left\{ \begin{array}{ll}
                            1 & \mbox{if $i = \argmax_j S\left(U_t, k_j\right)$},\\
                         0 & \mbox{otherwise.}
                       \end{array} \right.
\end{equation*}

We use two types of standard IR baselines: a TF-IDF~\cite{manning2008} and a BM25~\cite{robertson09} system. We also consider another IR baseline that employs an uncased pretrained BERT model as a static utterance encoder. In this baseline, we encode $U_t$ and each $k_i$ separately and then compute the cosine similarity between the pooled utterance outputs. 

\subsubsection{Neural Relevance Classification}
\label{sec:methods:task2:classification}

We also employ a BERT-based~\cite{Devlin2019BERTPO} neural model as a baseline knowledge selection system.
In particular, we train a binary classification model (Figure~\ref{fig:baseline_task2_bert}) over a pair of encoded texts as is done in prior Transformer sentence relationship models~\cite{Nogueira2019PassageRW}.
The model takes the concatenation of the utterances in $U_t$ and the sentences in $k_i$ as an input instance.
We use the final layer output $C$ at the same position to the $[CLS]$ token as input to a single layer feedforward network to obtain a probability $s_i$ that the $k_i$ is relevant to the given dialogue context $U_t$.

We finetune a pretrained BERT model using a binary cross-entropy loss as follows:
\begin{equation*}
  L = -\displaystyle\sum_{i\in I_{pos}} \log(s_i) - \displaystyle\sum_{i\in I_{neg}} \log(1 - s_i),
\end{equation*}
where $I_{pos}$ refers to the set of knowledges that are relevant for the given dialogue context and $I_{neg}$ refers to those that are not.

During training of the knowledge classifier, we experimented with sampling methods of negative knowledge candidates to be paired with a given dialogue context.
For dialogues annotated with domain-level knowledge, we chose negative candidates by sampling other documents in the same domain as the annotation.
For entity-level knowledge dialogues, we chose negative candidates by sampling other documents from the same entity as the provided annotation.
We built models in which the number of negative candidates for each positive example was varied from 1 to 13 in increments of 4 and found the best-performing model used $5$ negative candidates for each positive candidate.

\subsection{Knowledge-grounded Generation}
\label{sec:methods:task3}

In this section, we propose both extractive and generative approaches for the knowledge-grounded response generation task.

\subsubsection{Answer Extraction}
\label{sec:methods:task3:extraction}

The simplest method for knowledge-grounded response generation is to output a part of the selected knowledge snippets.
In this experiment, we developed an answer extraction baseline with the following heuristics:
\begin{itemize}
\item If multiple knowledge snippets are related to a given turn, randomly pick one of them. Otherwise, a sole snippet is taken as the source for answer extraction.
\item If the target snippet includes multiple paragraphs, extract only the first paragraph as a system response. Otherwise, the whole paragraph is considered as the output.
\end{itemize}

\subsubsection{Neural Response Generation}
\label{sec:methods:task3:neural}

\begin{figure}[t]
  \includegraphics[width=\linewidth]{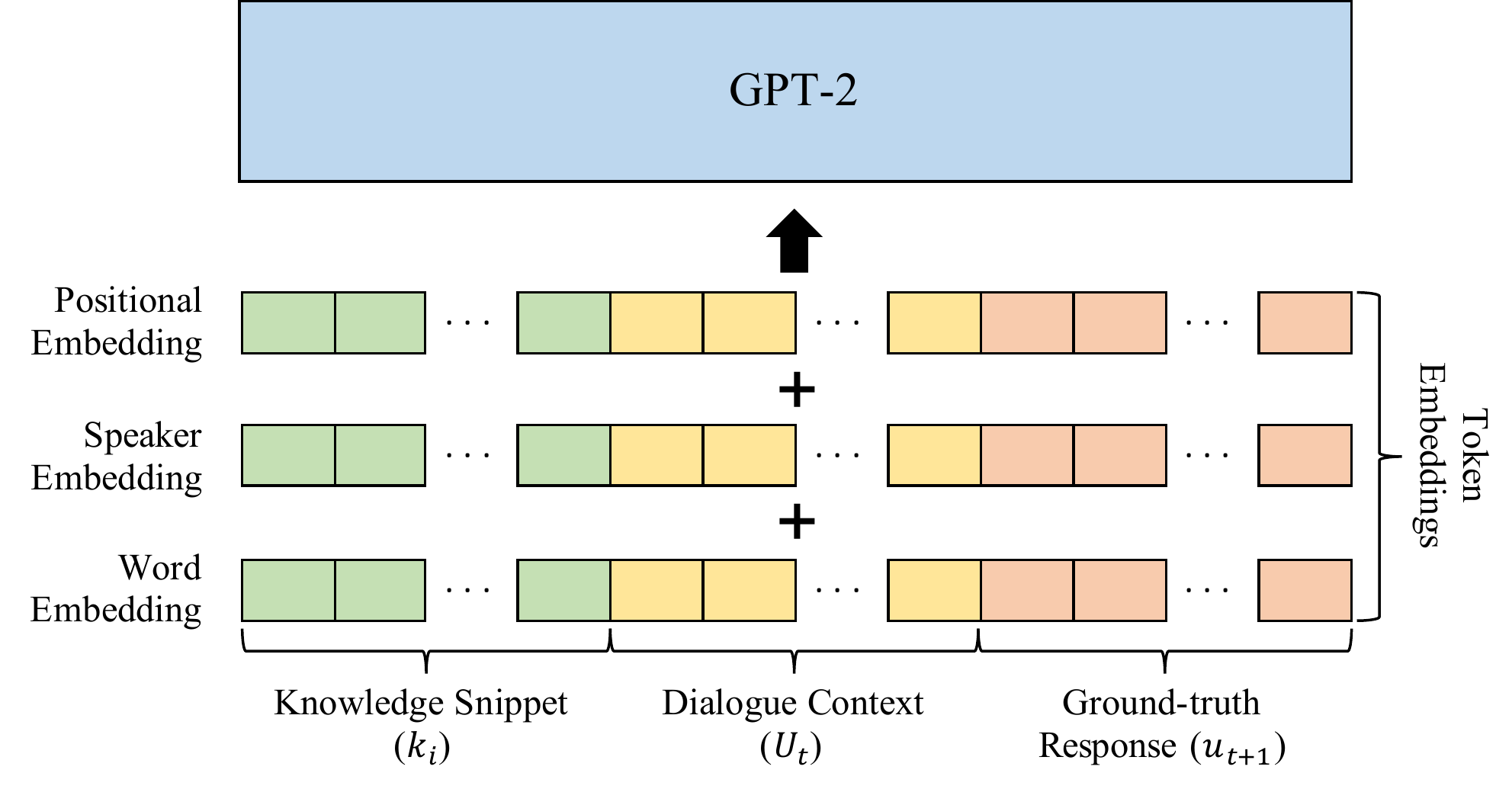}
  \caption{Input representation for GPT-2 w/ knowledge}
  \label{fig:gpt_input}
\end{figure}

Given the tremendous interest and success in leveraging large pre-trained language models for downstream NLP tasks in the community, our neural baseline leverages the Generative Pre-trained Transformer (GPT-2) model~\cite{radford2019language}.
We fine-tuned the GPT-2 \textit{small} model with a standard language modeling objective on our dataset, using both the knowledge-augmented and regular system turns as target sequences.
To show the influence of knowledge, we compared two variants of models with different inputs, as follows:
\begin{itemize}
  \item GPT-2 w/o knowledge: no knowledge was used during fine-tuning.
  \item GPT-2 w/ knowledge: the ground-truth knowledge snippets were concatenated to each input dialog context (Figure~\ref{fig:gpt_input}) for fine-tuning.
\end{itemize}

\begin{table}
  \centering
  \small
  \begin{tabular}{l c c c c}
    Method & Acc & P & R & F \\ \hline
    Anomaly Detection & 0.852 & 0.393 & 0.581 & 0.469 \\
    Classification & \textbf{0.998} & \textbf{0.989} & \textbf{0.994} & \textbf{0.991} \\
  \end{tabular}
  \caption{Comparisons of the knowledge-seeking turn detection performances between two baselines}
  \label{tbl:turn_detection_results}
\end{table}

We used the \textit{transformers} library~\cite{Wolf2019HuggingFacesTS}~\footnote{https://huggingface.co/transformers/} to fine-tune the models for a fixed number of 3 epochs with a truncation window of 256 tokens for both dialog context $U_t$ and knowledge snippet $k_i$.
We used a train batch size of 2, performed gradient accumulation every 8 steps and gradient clipping with a max norm of $1.0$, used the Adam optimizer and linearly decayed the learning rate from 6.25e-5 to 0 during fine-tuning.

We added special tokens for both speakers \textit{user} and \textit{agent} to our vocabulary, initialized their parameters randomly and learned them during fine-tuning.
We enriched the corresponding turns in the input with speaker embeddings at a token-level by identifying their token types, exactly as described in~\cite{wolf2019transfertransfo}.
We used top-$k$, top-$p$ nucleus sampling with temperature $T$~\cite{holtzman2019curious} for decoding, where $k=0$, $p=0.9$ and $T=0.7$.
We also set a maximum decode length of 40 tokens.

\section{Evaluation}
\label{sec:evaluation}

\subsection{Knowledge-seeking Turn Detection}
\label{sec:evaluation:task1}

First, we evaluated the knowledge-seeking turn detection performances of unsupervised anomaly detection (Section~\ref{sec:methods:task1:anomaly}) and supervised neural classification (Section~\ref{sec:methods:task2:classification}) methods.
Both models were built on all the user utterances in the training set and evaluated on the test set user turns in accuracy, precision, recall, and F-measure.

Table~\ref{tbl:turn_detection_results} shows that the unsupervised baseline has a limitation in distinguishing between API-based and knowledge-seeking turns, especially with many false positives.
On the other hand, the neural classifier achieved almost perfect performance in all the metrics.
Nevertheless, this utterance classifier may work well when restricted only to this data set or similar, due to lack of knowledge or API details incorporated into the model.
There is much room for improvement in making the model more generalizable to unseen domains or knowledge sources.

\begin{table}
  \centering
  \small
  \begin{tabular}{l c c c c}
    Method & & MRR@5 & R@1 & R@5  \\ \hline
    Retrieval (TF-IDF) & & 0.618 & 0.511 & 0.807 \\
    Retrieval (BM25) & & 0.611 & 0.498 & 0.827 \\
    Retrieval (BERT) & & 0.226 & 0.128 & 0.428 \\ \hdashline[.4pt/1pt]
    Classification (BERT) & & \textbf{0.891} & \textbf{0.834} & \textbf{0.976} \\
  \end{tabular}
  \caption{Comparisons of the knowledge selection performances by retrieval and classification methods}
  \label{tbl:ks_results}
\end{table}

\subsection{Knowledge Selection}
\label{sec:evaluation:task2}

Knowledge selection was evaluated using a number of standard IR metrics including recall (R@1 and R@5), and mean reciprocal rank (MRR@5).
For domain-knowledge dialogues, our total candidate set included all domain knowledges for the annotated domain, and for entity-knowledge dialogues our total candidate set included all entity knowledges for the annotated entity.

Table~\ref{tbl:ks_results} shows that our bag-of-words IR baselines (Section~\ref{sec:methods:task2:retrieval}) outperformed the static BERT encoder across all three metrics.
However, the neural classifier model (Section~\ref{sec:methods:task2:classification}) significantly outperformed the IR baselines, demonstrating the efficacy of downstream fine-tuning of large pre-trained neural representations.
That being said, there is still a substantial performance gap in the R@1 and MRR@5 metrics, leaving room for further research into knowledge selection on this data.

\begin{table*}
  \centering
  \small
  \begin{tabular}{l r r r r r r r r}
    Method & & PPL & Unigram F1 & Div. ($n$ = 1) & Div. ($n$ = 2) & BLEU-4 & METEOR & ROUGE-L \\ \hline
    Answer Extraction & & - & \bf{0.3215} & 0.0356 & 0.0892 & 0.0358 & 0.2543 & 0.1769 \\ \hdashline[.4pt/1pt]
    GPT-2 w/o knowledge & & 5.0906 & 0.2620 & 0.0509 & \bf{0.1589} & 0.0559 & 0.2202 & 0.1979 \\
    GPT-2 with knowledge & & \bf{4.1723} & 0.3175 & \bf{0.0509} & 0.1559 & \bf{0.0840} & \bf{0.2796} & \bf{0.2403} \\ \hdashline[.4pt/1pt]
    Human & & - & - & 0.0806 & 0.3055 & - & - & - \\
  \end{tabular}
  \caption{Automated evaluation results on knowledge-grounded response generation}
  \label{tbl:generation_automatic}
\end{table*}

\begin{table*}
  \centering
  \small
  \begin{tabular}{l l c r r r c r r r}
    & & & \multicolumn{3}{c}{Appropriateness} & & \multicolumn{3}{c}{Accuracy} \\
    Method & Baseline & & \%W & \%L & \%Tie & & \%W & \%L & \%Tie \\ \hline 
    Answer Extraction & Human & & 34.39 & 59.49 & 6.12 & & - & - & - \\
    GPT-2 w/o knowledge & Human & & - & - & - & & 4.59 & 27.76 & 67.65 \\
    GPT-2 with knowledge & Human & & 36.02 & 59.49 & 4.49 & & 5.31 & 22.96 & 71.74 \\ \hdashline[.4pt/1pt]
    GPT-2 with knowledge & Answer Extraction & & \textbf{56.33} & 31.02 & 12.65 & & - & - & - \\
    GPT-2 with knowledge & GPT-2 w/o knowledge & & - & - & - & & \textbf{22.55} & 17.04 & 60.41 \\
  \end{tabular}
  \caption{Human evaluation results on knowledge-grounded response generation}
  \label{tbl:generation_human}
\end{table*}

\subsection{Knowledge-grounded Generation}
\label{sec:evaluation:task3}

Responses by answer extraction (Section~\ref{sec:methods:task3:extraction}) and neural generation models (Section~\ref{sec:methods:task3:neural}) were first evaluated using the following automated metrics: perplexity, unigram F1, \textit{n}-gram diversity, BLEU-4, METEOR, and ROUGE-L.
The evaluation was done only on the augmented turns with the ground-truth knowledge, in order to characterize the models' ability to handle the external knowledge scenario.
Table~\ref{tbl:generation_automatic} shows that our generation models achieved better scores than the extractive baseline on most metrics.
Especially, the GPT-2 model with knowledge outperformed both the answer extraction baseline and the other GPT-2 variant with no knowledge in BLEU-4, METEOR, and ROUGE-L,
which indicates that our proposed neural model generates more human-like responses than the extractive baseline.

In addition, we also performed human evaluations of the generated responses with the following two crowdsourcing tasks:
\begin{itemize}
\item \textbf{Appropriateness}: given a dialogue context and a pair of responses generated by two methods, crowdworkers were asked to select a more appropriate response to the context.
\item \textbf{Accuracy}: given a knowledge snippet and a pair of responses generated by two methods, crowdworkers were asked to select a more accurate response to the knowledge.
\end{itemize}
In both tasks, we presented each instance to three crowdworkers; asked them to choose either response or `not sure' for the cases that are equally good or bad; and took the majority as the final label for the instance.
Table~\ref{tbl:generation_human} shows that our GPT-2 models generated more appropriate responses than the answer extraction baseline.
Comparing between two GPT-2 variants, the model with knowledge provided more accurate information based on explicitly given knowledge than the one without knowledge.
However, this accuracy gap between two models is not very big, which depicts the need to add more diversity in knowledge content which cannot be handled just by memorizing facts from the training data.

\section{Conclusions}
\label{sec:conclusions}
This paper proposed a new task-oriented conversational modeling problem grounded on unstructured domain knowledge, which aims to handle out-of-API coverage user requests.
To support research on our proposed tasks, we introduced an augmented version of MultiWOZ 2.1 dialogues with additional knowledge-seeking turns collected given external knowledge sources.
We presented baseline methods based both on non-machine learning approaches and neural model architectures.

Furthering this work, we plan to collect more dialogues including different domains, entities, and locales from the original ones for MultiWOZ 2.1.
Moreover, this new data set will include not only written conversations, but also spoken dialogues to evaluate the system performances for more realistic scenarios.
Then, all the data sets and the baselines will be released for establishing a new public benchmark in dialogue research.

In addition, we will continue to iterate on the models with the following potential enhancements:
end-to-end learning instead of the pipelined processing,
joint modeling of both knowledge-seeking and API-driven branches,
and few shot transfer learning for unseen domains or knowledge sources.

\bibliography{sigdial2020_kmdm}
\bibliographystyle{acl_natbib}

\clearpage

\appendix
\section{Appendices}

\subsection{Unstructured Knowledge Sources}
Figure~\ref{domain_knowledge} and Figure~\ref{entity_knowledge} show examples of knowledge snippets used in our data collection for domain- and entity-specific augmented turns, respectively.
While domain-level snippets include generic information that could be applicable over all the domain entities,
entity-level knowledge varies depending on a given entity even for the same question.

\begin{figure}[h]
  \small
  \begin{tabular}{l p{6cm}}
    \hline
    Domain & Hotel \\ \hdashline[.4pt/1pt]
    Title & How can I get an invoice? \\
    Body & The property can provide you with an invoice for your stay, so please contact them directly.\\ \hline
    Domain & Restaurant \\ \hdashline[.4pt/1pt]
    Title & Cancellation \\
    Body & You can cancel a reservation online or call the restaurant directly. Please note that some restaurants have implemented a 24-48 hour cancellation policy.\\ \hline
    Domain & Train \\ \hdashline[.4pt/1pt]
    Title & Discount Information for Children \\
    Body & One child ages 2-12 is eligible to receive a 50\% discount on the lowest available adult rail fare on most trains with each fare-paying adult (age 18+). \\ \hline
  \end{tabular}
  \caption{Examples of domain-level knowledge}
  \label{domain_knowledge}
\end{figure}

\begin{figure}[h]
  \small
  \begin{tabular}{l p{6cm}}
    \hline
    Domain & Hotel \\
    Entity & Gonville Hotel \\ \hdashline[.4pt/1pt]
    Title & What is the parking charge? \\
    Body &  Parking costs GBP 14 per day. \\ \hline
    Domain & Hotel \\
    Entity & Hamilton Lodge \\ \hdashline[.4pt/1pt]
    Title & Is there free parking at your lodge? \\
    Body & Hamilton Lodge offers free parking. \\ \hline
    Domain & Restaurants \\
    Entity & La Margherita \\ \hdashline[.4pt/1pt]
    Title & What credit cards are accepted? \\
    Body & La Margherita takes Visa, Mastercard and AMEX. \\ \hline
    Domain & Restaurants \\
    Entity & Peking Restaurant \\ \hdashline[.4pt/1pt]
    Title & What type of payments are accepted at this location? \\
    Body & Peking Restaurant accepts cash only. \\ \hline
  \end{tabular}
  \caption{Examples of entity-level knowledge}
  \label{entity_knowledge}
\end{figure}

\newpage
\subsection{Human Evaluation Tasks}

\begin{figure}[h]
  \includegraphics[width=\linewidth]{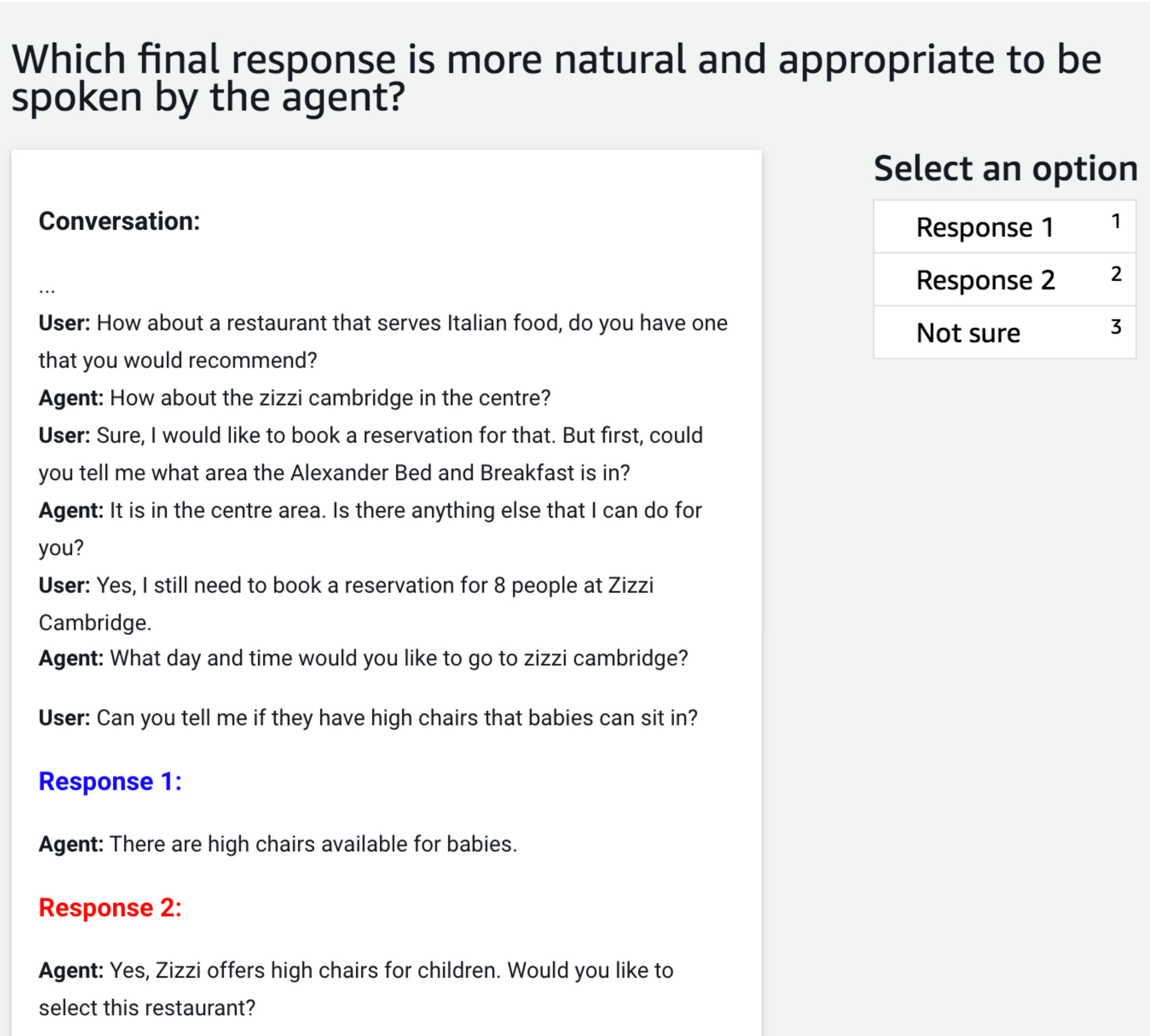}
  \caption{Crowdsourcing user interface for the human evaluation on appropriateness}
  \label{mturk_eval1}
\end{figure}

\begin{figure}[h]
  \includegraphics[width=\linewidth]{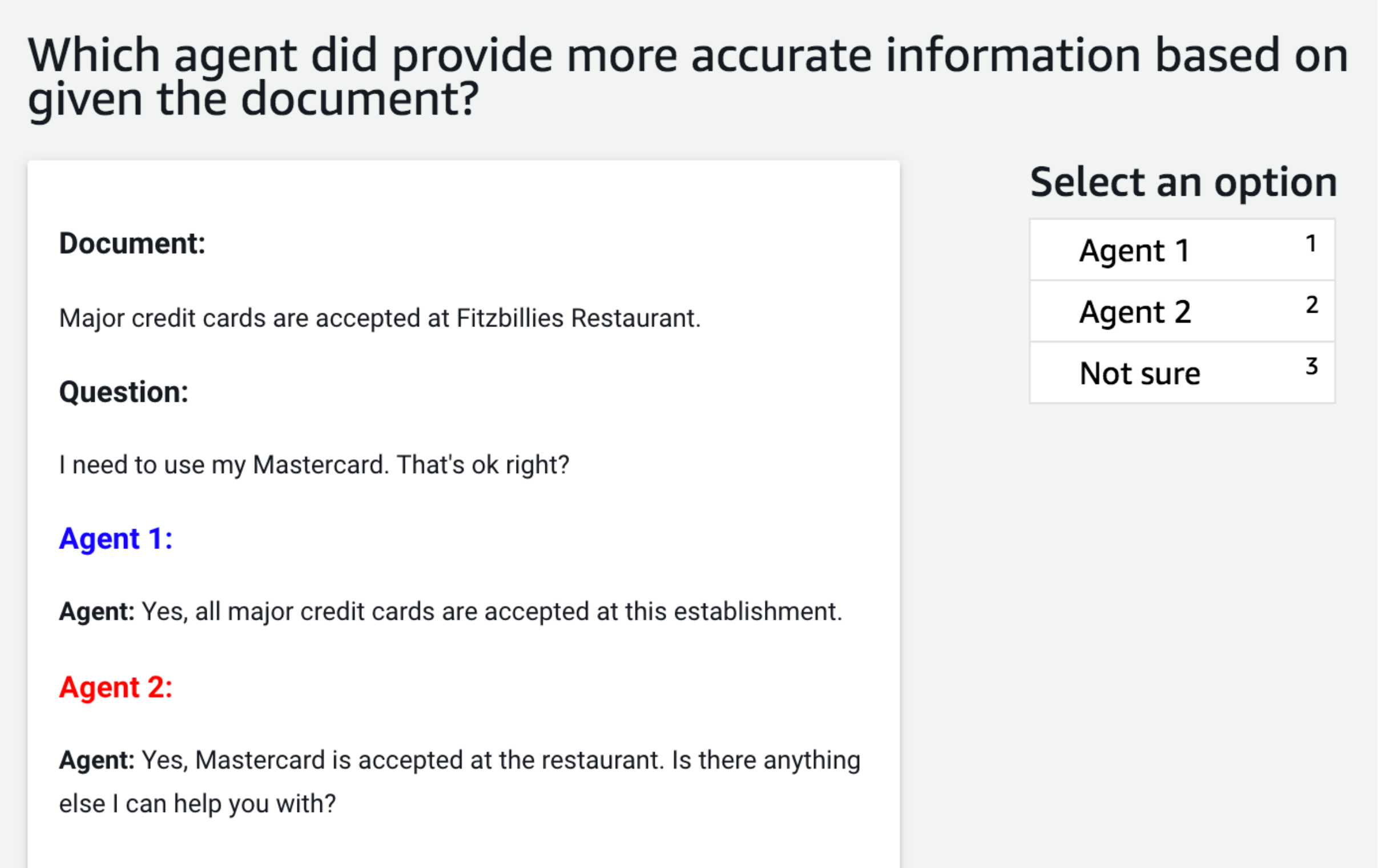}
  \caption{Crowdsourcing user interface for the human evaluation on accuracy}
  \label{mturk_eval2} 
\end{figure}

Figure~\ref{mturk_eval1} and Figure~\ref{mturk_eval2} show the crowdsourcing user interfaces for human evaluations on \textit{appropriateness} and \textit{accuracy}, respectively.

\subsection{Generated Responses}

Figure~\ref{fig:appendix_example} compares the example responses generated by the proposed methods.
GPT-2 models generated more natural and appropriate responses to be spoken by the agent than the answer extraction baseline.
Comparing between two GPT-2 variants, the model with knowledge provided more accurate information than the one without knowledge.
However, there is still room for improvement in both appropriateness and accuracy compared to the human responses.

\begin{figure*}[t]
  \small
  \begin{tabular}{l p{13.5cm}}
    \multicolumn{2}{c}{\bf{Example \#1}}\\
  \hline \multicolumn{2}{c}{Dialogue Context}\\ \hline
  User & I'm also looking for info on a hotel called allenbell.\\
  Agent & Would you like me to book it for you?\\
  User & Am I able to smoke while staying at this hotel?\\
  \hline \multicolumn{2}{c}{Relevant Knowledge}\\ \hline
       & Q: Can I smoke there?\\
       & A: Allenbell is a non-smoking establishment.\\
  \hline \multicolumn{2}{c}{Generated Responses}\\ \hline
  Human & The Allenbell does not allow smoking in their hotel. Can I help you answer any other questions?\\ \hdashline[.4pt/1pt]
  Extracted & Allenbell is a non-smoking establishment. \\ \hdashline[.4pt/1pt]
  GPT-2 w/o k. & Please contact us for more information \\ \hdashline[.4pt/1pt]
    GPT-2 with k. & No, they don't allow smoking in the hotel. Would you like to try a different hotel? \\ \hline
    \\
    \multicolumn{2}{c}{\bf{Example \#2}}\\
  \hline \multicolumn{2}{c}{Dialogue Context}\\ \hline  
  User & Great! I'm also looking for a place to eat in the same area. Something not too expensive, but not cheap.\\
  Agent & There are 21 places in the centre of town. What type of cuisine would you like?\\
  User & I am interested in moderately priced international food.\\
  Agent & I'd recommend the Varsity Restaurant. Would you like to book a table? \\
  \hline \multicolumn{2}{c}{Relevant Knowledge}\\ \hline
       & Q: Payment options \\
       & A: We prefer cash payment. We accept MasterCard and Visa with a \$20 minimum purchase.\\
  \hline \multicolumn{2}{c}{Generated Responses}\\ \hline
  Human & Yes. Cash is actually the preferred payment method. \\ \hdashline[.4pt/1pt]
  Extracted & We prefer cash payment. We accept MasterCard and Visa with a \$20 minimum purchase. \\ \hdashline[.4pt/1pt]
  GPT-2 w/o k. & Yes, all major credit cards are accepted. Would you like me to make a reservation for you? \\ \hdashline[.4pt/1pt]
    GPT-2 with K. & Yes, the restaurant accepts cash. Do you need anything else? \\ \hline
    \\
    \multicolumn{2}{c}{\bf{Example \#3}}\\
    \hline \multicolumn{2}{c}{Dialogue Context}\\ \hline  
User & Can you help me find a restaurant that serves African food with a moderate price range please \\
Agent & I am sorry, there aren't any options available. May I ask if there is another type of restaurant you would be interested in? \\
User & Yes how about Asian food in the same price range? \\
Agent & Yes I have the Yippee Noodle Bar in the center of town on King street in the moderate price range. They serve Asian cuisine. Is there anything else I can do for you? \\
User & Will you please book a table for 6 at 13:45 on Friday? \\
Agent & Booking was successful. The table will be reserved for 15 minutes.
Reference number is : 3GLLHSNN.  \\
    \hline \multicolumn{2}{c}{Relevant Knowledge}\\ \hline
       & Q: What parking options are available at your location? \\
       & A: Only street parking is available. \\
    \hline \multicolumn{2}{c}{Generated Responses}\\ \hline
    Human & There is only parking on the street. Anything else I can do for you? \\ \hdashline[.4pt/1pt]
    Extracted & Only street parking is available.\\ \hdashline[.4pt/1pt]
    GPT-2 w/o k. & No, the restaurant does not offer parking for parking. Would you like me to book a reservation for you? \\ \hdashline[.4pt/1pt]
    GPT-2 with k. & Yes it does. Would you like me to make reservations for you? \\ \hline
\end{tabular}
\caption{Example responses generated by our proposed methods}
\label{fig:appendix_example}
\end{figure*}

\end{document}